\pdfoutput=1

\documentclass[11pt]{article}

\usepackage[]{EMNLP2023}

\usepackage{times}
\usepackage{latexsym}
\usepackage{arydshln}

\usepackage[T1]{fontenc}

\usepackage[utf8]{inputenc}

\usepackage{microtype}

\usepackage{inconsolata}
\usepackage{xcolor}

\title{Human-like Summarization Evaluation with ChatGPT}

\author{Mingqi Gao, Jie Ruan, Renliang Sun, Xunjian Yin, Shiping Yang, Xiaojun Wan\\
         Wangxuan Institute of Computer Technology, Peking University\\
         \texttt{\{gaomingqi, xjyin, wanxiaojun\}@pku.edu.cn}\\
         \texttt{\{ruanjie, sunrenliang\}@stu.pku.edu.cn}\\ 
         \texttt{yangshiping@bupt.edu.cn}}

\begin{document}
\maketitle
\begin{abstract}
Evaluating text summarization is a challenging problem, and existing evaluation metrics are far from satisfactory. In this study, we explored ChatGPT's ability to perform human-like summarization evaluation using four human evaluation methods on five datasets. We found that ChatGPT was able to complete annotations relatively smoothly using Likert scale scoring, pairwise comparison, Pyramid, and binary factuality evaluation. Additionally, it outperformed commonly used automatic evaluation metrics on some datasets. Furthermore, we discussed the impact of different prompts, compared its performance with that of human evaluation, and analyzed the generated explanations and invalid responses.
\end{abstract}

\section{Introduction}

Text summarization is a task that involves generating a condensed version of one or multiple documents. Thanks to the advancements in deep learning-based techniques, automatic summarization has made significant strides. Specifically, the emergence of large language models such as InstructGPT has resulted in comparable performance to reference summaries written by humans, even in zero-shot settings \citep{zhang2023benchmarking}.

Evaluating text summarization, like other text generation tasks, is a challenging problem. While human evaluation is considered the gold standard, it is expensive and time-consuming. As a result, automatic evaluation metrics play a crucial role. ROUGE \citep{lin-2004-rouge} and its variants, which are based on reference summaries and n-gram matching, are widely accepted and used in various types of summarization. However, surface-level word matching cannot accurately reflect the quality of the summary. Additionally, it is challenging to evaluate the factual accuracy of the summary without utilizing the source document. Recently, evaluation metrics based on pre-trained models such as BERTScore \citep{zhang2020bertscore} and BARTScore \citep{yuan2021bartscore} have achieved better correlation with human judgments. Factuality evaluation methods based on entailment classification, such as FactCC \citep{kryscinski-etal-2020-evaluating}, and question answering, such as FEQA \citep{durmus-etal-2020-feqa}, have also been used to evaluate the factual consistency of summaries. Despite the existence of advanced automatic evaluation metrics, their performance, usability, and interpretability are still far from satisfactory.

Large language models (LLMs) offer completely different possibilities for the automatic evaluation of summarization. GPT-3 \citep{brown2020language} has the ability of in-context learning, and instruction tuning allows LLMs to align with human evaluation \citep{ouyang2022training}. These two abilities make it possible for LLMs to mimic the behavior of human evaluators, who generally evaluate summaries by understanding examples and instructions. We refer to this automatic evaluation method that views large models as human evaluators as human-like automatic evaluation. The most prominent feature of this evaluation method is its flexibility, which unifies all types of automatic evaluation in form and can simulate many of the practices of human evaluators. Unlike previous automatic evaluation metrics that give one or more numerical values as evaluation results, the evaluation results of this human-like automatic evaluation are fully reflected in the generated responses, which may include scoring, comparison, labels, and explanations.

We conducted an evaluation of the evaluation ability of ChatGPT, a recently popular LLM, using four commonly used human evaluation methods for summarization. The methods include Likert scale scoring, pairwise comparison, Pyramid \citep{nenkova-passonneau-2004-evaluating}, and binary factuality evaluation. Our findings indicate that ChatGPT is capable of completing annotations relatively smoothly using these methods. In addition, our results demonstrate that ChatGPT outperforms commonly used automatic evaluation metrics on some datasets. Furthermore, we analyzed the impact of different prompts, compared the performance of ChatGPT with human evaluation, and examined the quality of the generated explanations and invalid responses.

\section{Preliminary}

\subsection{Automatic Evaluation Metrics}
We select several evaluation metrics that are commonly used in summarization:

\textbf{ROUGE} \citep{lin-2004-rouge}, which is the dominant automatic evaluation metric in summarization, is widely used by researchers. The most commonly used ROUGE measures are ROUGE-1, ROUGE-2, and ROUGE-L, which evaluate the similarity between two texts based on the overlap of unigrams, bigrams, and the longest common sequence.

\textbf{BERTScore} \citep{zhang2020bertscore} assesses the similarity between two texts at the token level by measuring the soft overlap using contextual embeddings from BERT. Similarly, \textbf{MoverScore} \citep{zhao-etal-2019-moverscore} uses n-gram embeddings that are pooled from BERT to compute the semantic distance between two texts at the n-gram level.

\textbf{BARTScore} \citep{yuan2021bartscore} \footnote{ \url{https://github.com/neulab/BARTScore}, also for ROUGE, BERTScore, and MoverScore.} views evaluation as a natural language generation task and considers that when the quality of the generated text is higher, BART is more likely to generate it from the source text or the reference, or to generate the reference from it. BARTScore can be flexibly applied to evaluate text from various perspectives.

\textbf{FactCC}\footnote{  \url{https://github.com/salesforce/factCC}} and \textbf{DAE} \footnote{ \url{https://github.com/tagoyal/factuality-datasets}} are two factuality metrics based on classification. When evaluating a summary, we use NLTK \footnote{version 3.7, \url{https://www.nltk.org/}} to split it into individual sentences and classify each one as factually correct or not. The factual score of the summary is then calculated as the ratio of sentences that are factually correct.

\subsection{Human Evaluation Methods}
There are several commonly used methods for human evaluation, including the Likert scale scoring and pairwise comparison for general text generation, as well as Pyramid and binary factuality evaluation specifically designed for summarization. After introducing each method, we will list the datasets we used that were annotated in this way.

\textbf{Likert scale scoring} is the most common method for human evaluation. Specifically, given a source document and a generated summary, annotators rate the summary on several dimensions. Typically, this is an absolute evaluation, meaning each summary is evaluated individually without explicit comparison to other summaries. Dimensions usually include factual consistency, informativeness, fluency, etc. The rating scale is usually 1 (worst) to 5 (best). We used SummEval \citep{fabbri-etal-2021-summeval} and Newsroom datasets \citep{grusky-etal-2018-newsroom}.

\textbf{Pairwise comparison} is a relative human evaluation method. Given a source document and two generated summaries, annotators choose the one that is of higher quality. This method is used in reinforcement learning based human feedback for summarization. We used the TLDR dataset \citep{stiennon2022learning}.

\textbf{Pyramid} \citep{nenkova-passonneau-2004-evaluating} is a human evaluation method designed for summarization that is based on reference summaries. Prior to human annotation, several semantic content units (SCUs) are extracted from the reference summary. For each SCU, annotators judge whether it presents in the generated summary. For single-document summarization, the final score of the summary is the proportion of SCUs it contains. We used the REALSumm dataset \citep{bhandari-etal-2020-evaluating}.

\textbf{Binary factuality evaluation} is a method for evaluating the factual correctness of summaries. Given a source document and a sentence in the generated summary, annotators judge whether the sentence is faithful to the source document. We used the QAGS dataset \citep{wang-etal-2020-asking}.

\section{Experiments}

\subsection{Model and Parameters}
We used the ChatGPT API (gpt-3.5-turbo-0301) provided by OpenAI for our experiments. To reduce randomness, we set \texttt{temperature} to 0. In addition, we set \texttt{max\_tokens} to 256. We kept the default values for other parameters.

\subsection{Prompt Design}

When designing prompts, we made it as identical as possible to the original instructions of human evaluations.

\begin{figure}[h]
    \centering
\fcolorbox{black}{gray!10}{\parbox{0.9\linewidth}{
Evaluate the quality of summaries written for a news article. Rate each summary on four dimensions: \textcolor{blue}{\{Dimension\_1\}}, \textcolor{blue}{\{Dimension\_2\}}, \textcolor{blue}{\{Dimension\_3\}}, and \textcolor{blue}{\{Dimension\_4\}}. You should rate on a scale from 1 (worst) to 5 (best). \\

Article: \textcolor{blue}{\{Article\}} 

Summary: \textcolor{blue}{\{Summary\}}}}
    \caption{The template for Likert scale scoring.}
    \label{fig:likert_scale}
\end{figure}




Figure \ref{fig:likert_scale} shows the template for Likert scale scoring. ChatGPT is asked to rate four dimensions at a time. For SummEval, the four dimensions are relevance, faithfulness \footnote{The original term used in SummEval was "consistency". Since we did not add definitions of the dimensions in the prompt, we used "faithfulness", which is more representative of its actual meaning}, fluency, and coherence. For Newsroom, the four dimensions are relevance, informativeness, fluency, and coherence. Figure \ref{fig:pairwise} shows the template for pairwise comparison.

\begin{figure}[h]
    \centering
    \fcolorbox{black}{gray!10}{\parbox{0.9\linewidth}{
    
    Given a new article, which summary is better? Answer "Summary 0" or "Summary 1". You do not need to explain the reason.\\
    
    Article: \textcolor{blue}{\{Article\}} 
    
    Summary 0: \textcolor{blue}{\{Summary\_0\}}
    
    Summary 1: \textcolor{blue}{\{Summary\_1\}}}}
    \caption{The template for pairwise comparison.}
    \label{fig:pairwise}
\end{figure}

Figure \ref{fig:pyramid} shows the template for Pyramid. The number of SCUs depends on the content of the reference summary, up to 16.

\begin{figure}[h]
    \centering
    \fcolorbox{black}{gray!10}{\parbox{0.9\linewidth}{
    
    You are given a summary and some semantic content units. For each semantic unit, mark "Yes" if it can be inferred from the summary, otherwise mark "No".\\

    Summary: \textcolor{blue}{\{Summary\}}

    Semantic content units:
    
    1. \textcolor{blue}{\{SCU\_1\}}

    2. \textcolor{blue}{\{SCU\_2\}}

    ......

    n. \textcolor{blue}{\{SCU\_n\}}
}}
    \caption{The template for Pyramid.}
    \label{fig:pyramid}
\end{figure}

Figure \ref{fig:binary_fact} shows the template for binary factuality evaluation. The sentences are from the generated summaries.

\begin{figure}[h]
    \centering
    \fcolorbox{black}{gray!10}{\parbox{0.9\linewidth}{
    
    Is the sentence supported by the article? Answer "Yes" or "No".\\

    Article: \textcolor{blue}{\{Article\}}
    
    Sentence: \textcolor{blue}{\{Sentence\}}
}}
    \caption{The template for binary factuality evaluation.}
    \label{fig:binary_fact}
\end{figure}

\subsection{Post-processing of Results}
The vast majority of ChatGPT responses contained annotation results, which can be extracted by some simple rules. For invalid responses, we considered them as failing to complete the tagging successfully and marked them as NAN (not a number).

\subsection{Evaluation}

For Likert scale scoring, we computed sample-level, system-level, and dataset-level correlation with human judgments. For the other human evaluation methods, we calculated the accuracy of the responses generated by ChatGPT using human annotation as the answer.

\subsection{Results}

Tables \ref{tab:summeval} and \ref{tab:newsroom} show that ChatGPT has a good ability to evaluate summaries with Likert scale scoring. On SummEval, it performs substantially better than the existing evaluation metrics. On Newsroom, it performs second only to BARTScore\_s\_h and BARTScore\_cnn\_s\_h.

Tables \ref{tab:tldr}, \ref{tab:realsumm} and \ref{tab:qags} illustrate that ChatGPT can also perform relatively smoothly on pairwise comparisons, Pyramid, and binary factuality evaluation. Nevertheless, from the current experimental results, ChatGPT has not yet shown a very large advantage except on QAGS\_XSUM.

\begin{table*}[]
\scriptsize
\begin{tabular}{l|rrr|rrr|rrr|rrr}
\hline
 & \multicolumn{3}{c|}{consistency} & \multicolumn{3}{c|}{relevance} & \multicolumn{3}{c|}{fluency} & \multicolumn{3}{c}{coherence} \\ \cline{2-13} 
Metric Name & \multicolumn{1}{c}{sample} & \multicolumn{1}{c}{system} & \multicolumn{1}{c|}{dataset} & \multicolumn{1}{c}{sample} & \multicolumn{1}{c}{system} & \multicolumn{1}{c|}{dataset} & \multicolumn{1}{c}{sample} & \multicolumn{1}{c}{system} & \multicolumn{1}{c|}{dataset} & \multicolumn{1}{c}{sample} & \multicolumn{1}{c}{system} & \multicolumn{1}{c}{dataset} \\ \hline
ROUGE-1 & 0.153 & 0.744 & 0.137 & 0.326 & 0.744 & 0.302 & 0.113 & 0.730 & 0.080 & 0.167 & 0.506 & 0.184 \\
ROUGE-2 & 0.179 & 0.779 & 0.129 & 0.290 & 0.621 & 0.245 & 0.156 & 0.690 & 0.062 & 0.184 & 0.335 & 0.145 \\
ROUGE-L & 0.111 & 0.112 & 0.109 & 0.311 & 0.362 & 0.284 & 0.103 & 0.306 & 0.079 & 0.128 & 0.138 & 0.141 \\
BERTScore & 0.105 & -0.077 & 0.118 & 0.312 & 0.324 & 0.362 & 0.189 & 0.246 & 0.150 & 0.284 & 0.477 & 0.317 \\
MoverScore & 0.151 & 0.679 & 0.150 & 0.318 & 0.724 & 0.294 & 0.126 & 0.687 & 0.119 & 0.159 & 0.474 & 0.178 \\
BARTScore\_s\_h & 0.299 & 0.800 & 0.269 & 0.264 & 0.524 & 0.363 & 0.243 & 0.614 & 0.187 & 0.322 & 0.477 & 0.335 \\
BARTScore\_h\_r & 0.097 & 0.606 & 0.101 & 0.178 & 0.147 & 0.246 & 0.002 & 0.261 & 0.000 & 0.017 & -0.115 & 0.064 \\
BARTScore\_r\_h & -0.075 & -0.556 & -0.090 & -0.081 & -0.112 & -0.136 & 0.013 & -0.212 & 0.019 & 0.044 & 0.165 & -0.010 \\
BARTScore\_cnn\_s\_h & 0.367 & 0.435 & 0.334 & 0.356 & 0.765 & 0.394 & 0.349 & 0.746 & 0.285 & 0.448 & 0.700 & 0.408 \\
BARTScore\_cnn\_h\_r & 0.171 & 0.771 & 0.106 & 0.320 & 0.456 & 0.244 & 0.111 & 0.561 & 0.066 & 0.153 & 0.174 & 0.130 \\
BARTScore\_cnn\_r\_h & 0.001 & -0.079 & -0.004 & 0.146 & 0.312 & 0.221 & 0.107 & 0.297 & 0.145 & 0.228 & 0.506 & 0.236 \\
ChatGPT & \textbf{0.435} & \textbf{0.833} & \textbf{0.425} & \textbf{0.433} & \textbf{0.901} & \textbf{0.445} & \textbf{0.419} & \textbf{0.889} & \textbf{0.410} & \textbf{0.561} & \textbf{0.832} & \textbf{0.557} \\ \hline
\end{tabular}
\caption{Spearman's $\rho$ of sample level, system level, and dataset level on SummEval. }
\label{tab:summeval}
\end{table*}

\begin{table*}[]
\scriptsize
\begin{tabular}{l|rrr|rrr|rrr|rrr}
\hline
 & \multicolumn{3}{c|}{coherence} & \multicolumn{3}{c|}{fluency} & \multicolumn{3}{c|}{informativeness} & \multicolumn{3}{c}{relevance} \\ \cline{2-13} 
Metric Name & \multicolumn{1}{c}{sample} & \multicolumn{1}{c}{system} & \multicolumn{1}{c|}{dataset} & \multicolumn{1}{c}{sample} & \multicolumn{1}{c}{system} & \multicolumn{1}{c|}{dataset} & \multicolumn{1}{c}{sample} & \multicolumn{1}{c}{system} & \multicolumn{1}{c|}{dataset} & \multicolumn{1}{c}{sample} & \multicolumn{1}{c}{system} & \multicolumn{1}{c}{dataset} \\ \hline
ROUGE-1 & 0.095 & 0.429 & 0.100 & 0.104 & 0.429 & 0.064 & 0.130 & 0.286 & 0.149 & 0.147 & 0.357 & 0.122 \\
ROUGE-2 & 0.025 & 0.321 & 0.080 & 0.047 & 0.321 & 0.045 & 0.078 & 0.250 & 0.158 & 0.090 & 0.357 & 0.124 \\
ROUGE-L & 0.064 & 0.357 & 0.079 & 0.072 & 0.357 & 0.045 & 0.089 & 0.214 & 0.137 & 0.106 & 0.321 & 0.101 \\
BERTScore & 0.148 & 0.429 & 0.169 & 0.170 & 0.429 & 0.154 & 0.131 & 0.286 & 0.196 & 0.163 & 0.357 & 0.176 \\
MoverScore & 0.162 & 0.429 & 0.173 & 0.120 & 0.429 & 0.112 & 0.188 & 0.286 & 0.232 & 0.195 & 0.357 & 0.192 \\
BARTScore\_s\_h & \textbf{0.679} & \textbf{0.964} & \textbf{0.656} & \textbf{0.670} & \textbf{0.964} & \textbf{0.615} & \textbf{0.646} & \textbf{0.821} & \textbf{0.645} & \textbf{0.604} & \textbf{0.893} & \textbf{0.588} \\
BARTScore\_h\_r & 0.329 & 0.286 & 0.302 & 0.292 & 0.286 & 0.261 & 0.419 & 0.429 & 0.386 & 0.363 & 0.357 & 0.386 \\
BARTScore\_r\_h & -0.311 & -0.571 & -0.249 & -0.215 & -0.571 & -0.232 & -0.423 & -0.750 & -0.346 & -0.334 & -0.607 & -0.305 \\
BARTScore\_cnn\_s\_h & 0.653 & 0.893 & 0.623 & 0.640 & 0.893 & 0.596 & 0.616 & 0.750 & 0.592 & 0.567 & 0.786 & 0.557 \\
BARTScore\_cnn\_h\_r & 0.239 & 0.429 & 0.215 & 0.235 & 0.429 & 0.165 & 0.284 & 0.429 & 0.239 & 0.267 & 0.464 & 0.221 \\
BARTScore\_cnn\_r\_h & 0.316 & 0.429 & 0.333 & 0.353 & 0.429 & 0.330 & 0.242 & 0.286 & 0.289 & 0.245 & 0.357 & 0.292 \\
ChatGPT & 0.484 & 0.821 & 0.476 & 0.480 & 0.607 & 0.471 & 0.521 & 0.607 & 0.508 & 0.524 & 0.714 & 0.521 \\ \hline
\end{tabular}
\caption{Spearman's $\rho$ of sample level, system level, and dataset level on Newsroom. }
\label{tab:newsroom}
\end{table*}

\begin{table}[]
\begin{tabular}{lr}
\hline
Metric Name & \multicolumn{1}{c}{Accuracy} \\ \hline
ROUGE-1 & 0.5869 \\
ROUGE-2\_f & 0.4997 \\
ROUGE-L\_f & 0.5647 \\
BARTScore & 0.5674 \\
MoverScore & 0.5864 \\
BARTScore\_s\_h & 0.5858 \\
BARTScore\_h\_r & 0.6151 \\
BARTScore\_r\_h & 0.5317 \\
BARTScore\_cnn\_s\_h & 0.5880 \\
BARTScore\_cnn\_h\_r & 0.5934 \\
BARTScore\_cnn\_r\_h & 0.5089 \\
ChatGPT & \textbf{0.6178} \\ \hline
\end{tabular}
\caption{Accuracy of pairwise comparison on TLDR. }
\label{tab:tldr}
\end{table}

\begin{table}[]
\begin{tabular}{lr}
\hline
 Metric Name & \multicolumn{1}{c}{Accuracy} \\ \hline
DAE & 0.6304 \\
FactCC & 0.5362 \\
ChatGPT & \textbf{0.6436} \\ \hline
\end{tabular}
\caption{Accuracy of the binary determination of SCUs on REALSumm. }
\label{tab:realsumm}
\end{table}

\begin{table}[]
\begin{tabular}{lrr}
\hline
 & \multicolumn{1}{c}{QAGS\_CNN} & \multicolumn{1}{c}{QAGS\_XSUM} \\ \hline
DAE & 0.8459 & 0.6360 \\
FactCC & 0.7731 & 0.4937 \\
ChatGPT & \textbf{0.8488} & \textbf{0.7573} \\ \hline
\end{tabular}
\caption{Accuracy of binary factuality evaluation on QAGS. }
\label{tab:qags}
\end{table}

\section{Analysis and Discussion}

\subsection{Impact of different prompts}

\begin{table*}[]
\scriptsize
\begin{tabular}{l|rrr|rrr|rrr|rrr}
\hline
 & \multicolumn{3}{c|}{consistency} & \multicolumn{3}{c|}{relevance} & \multicolumn{3}{c|}{fluency} & \multicolumn{3}{c}{coherence} \\ \cline{2-13} 
 & \multicolumn{1}{c}{sample} & \multicolumn{1}{c}{system} & \multicolumn{1}{c|}{dataset} & \multicolumn{1}{c}{sample} & \multicolumn{1}{c}{system} & \multicolumn{1}{c|}{dataset} & \multicolumn{1}{c}{sample} & \multicolumn{1}{c}{system} & \multicolumn{1}{c|}{dataset} & \multicolumn{1}{c}{sample} & \multicolumn{1}{c}{system} & \multicolumn{1}{c}{dataset} \\ \hline
ChatGPT & 0.435 & 0.833 & 0.425 & 0.433 & 0.901 & 0.445 & 0.419 & 0.889 & 0.410 & 0.561 & 0.832 & 0.557 \\
ChatGPT+def & 0.471 & 0.786 & 0.479 & 0.453 & 0.877 & 0.479 & 0.347 & 0.606 & 0.341 & 0.568 & 0.802 & 0.570 \\
ChatGPT+def+ins & 0.338 & -0.149 & 0.302 & 0.396 & -0.079 & 0.433 & 0.349 & 0.016 & 0.325 & 0.501 & 0.338 & 0.494 \\
ChatGPT+sys\_prompt & 0.414 & 0.007 & 0.376 & 0.334 & 0.268 & 0.365 & 0.390 & 0.149 & 0.362 & 0.473 & 0.552 & 0.470 \\
\hdashline[0.5pt/5pt]
Annotator\_0 & \textbf{0.843} & \textbf{0.990} & \textbf{0.902} & 0.748 & \textbf{0.968} & 0.816 & 0.740 & \textbf{0.960} & 0.775 & 0.845 & 0.929 & 0.884 \\
Annotator\_1 & 0.813 & 0.965 & 0.881 & \textbf{0.767} & 0.953 & \textbf{0.823} & \textbf{0.847} & 0.843 & \textbf{0.876} & \textbf{0.889} & \textbf{0.982} & \textbf{0.913} \\
Annotator\_2 & 0.712 & 0.973 & 0.797 & 0.743 & 0.944 & 0.747 & 0.613 & 0.923 & 0.700 & 0.790 & 0.932 & 0.820 \\ \hline
\end{tabular}

\caption{Spearman's $\rho$ of sample level, system level, and dataset level on SummEval. Annotator\_0, Annotator\_1, Annotator\_2 refer to the three expert annotators. We compute the correlation coefficient between the score given by a particular annotator and the average score of the three. "+def" means adding dimension definitions in the prompt. "+ins" means adding step instructions in the prompt. Please see the example in Figure \ref{fig:likert_scale_detail} for dimension definitions and step instructions. "+sys\_prompt" denotes setting system prompt.}
\label{tab:summeval_detail}
\end{table*}

\begin{figure*}[h]
\centering
\fcolorbox{black}{gray!10}{\parbox{\linewidth}{

Imagine you are a human annotator now. You will evaluate the quality of summaries written for a news article. \color{red}{Please follow these steps:\\

1. Carefully read the news article, and be aware of the information it contains.

2. Read the proposed summary.

3. Rate the summary on four dimensions: relevance, consistency, fluency, and coherence. You should rate on a scale from 1 (worst) to 5 (best).}\\

\color{orange}{Definitions are as follows:

Relevance: The rating measures how well the summary captures the key points of the article. Consider whether all and only the important aspects are contained in the summary.

Consistency: The rating measures whether the facts in the summary are consistent with the facts in the original article. Consider whether the summary does reproduce all facts accurately and does not make up untrue information.

Fluency: This rating measures the quality of individual sentences, whether they are well-written and grammatically correct. Consider the quality of individual sentences.

Coherence: The rating measures the quality of all sentences collectively, to fit together and sound natural. Consider the quality of the summary as a whole.}\\

\color{black}{The article and the summary are given below:

Article:} \color{blue}{\{Article\}} 

\color{black}{Summary:} \color{blue}{\{Summary\}}}}

    \caption{The template for Likert scale scoring with step instructions (in \textcolor{red}{red}) and dimension definitions (in \textcolor{orange}{orange}) on SummEval.}
    \label{fig:likert_scale_detail}
\end{figure*}

\begin{table*}[]
\begin{tabular}{|c|l|}
\hline
\textbf{Summary} & \begin{tabular}[c]{@{}l@{}}Paul Merson was brought on with only seven minutes remaining in his team's \\ 0-0 draw with Burnley. Andros Townsend scored the Tottenham midfielder in \\ the 89th minute. Paul Merson had another dig at Andros Townsend after his \\ appearance. The midfielder had been brought on to the England squad last \\ week. Click here for all the latest arsenal news."\end{tabular} \\ \hline
\textbf{Expert annotator} & \begin{tabular}[c]{@{}l@{}}Relevance: 1.67 (2, 1, 2)\\ Consistency (Faithfulness): 1.00 (1, 1, 1)\\ Fluency: 3.00 (4, 2, 3)\\ Coherence: 1.33 (2, 1, 1)\end{tabular} \\ \hline
\textbf{ChatGPT} & \begin{tabular}[c]{@{}l@{}}Relevance: 2 \\ Faithfulness: 2\\ Fluency: 2\\ Coherence: 2\\ \\ Overall score: 2\\ \\ This summary is not very relevant to the article as it only mentions a few \\ details and leaves out important information. It is not very faithful as \textcolor{red}{it}\\ \textcolor{red}{ misrepresents the article by suggesting that Andros Townsend scored in the} \\ \textcolor{red}{Tottenham game, when in fact he was only brought on as a substitute.} \textcolor{blue}{The} \\ \textcolor{blue}{fluency and coherence} are also poor, with awkward phrasing and lack of \\ logical flow.\end{tabular} \\ \hline
\textbf{ChatGPT+def} & \begin{tabular}[c]{@{}l@{}}Relevance: 2\\ Consistency: 1\\ Fluency: 2\\ Coherence: 1\\     \\ Explanation: \\ The summary is not relevant as it does not capture the key points of the article. \\ It is inconsistent with the facts in the original article, \textcolor{red}{as it incorrectly states} \\ \textcolor{red}{that Andros Townsend scored in the 89th minute, when in fact he was brought} \\ \textcolor{red}{on in the 83rd minute.} \textcolor{blue}{The fluency} of the summary is poor, with grammatical \\ errors and incomplete sentences. \textcolor{blue}{The coherence} of the summary is also poor,\\ as it does not flow well and does not make sense as a whole.\end{tabular} \\ \hline
\end{tabular}
\caption{Examples of ratings and explanations generated by ChatGPT. The \textcolor{red}{red} parts show the factual errors of the summary. The \textcolor{blue}{blue} parts show the explanations for fluency and coherence. The long source document is not shown. “Expert Annotator” shows the average of scores given by three expert annotators, with each person's score in parentheses.}
\label{tab:expl_cases}
\end{table*}

We tried several different prompts on SummEval. As shown in Figure \ref{fig:likert_scale_detail}, more detailed \textbf{step instructions} and \textbf{dimension definitions} are added. These instructions and definitions are from the original human evaluation. In addition, we consider setting the system prompt as "You are a human annotator that rates the quality of summaries." when using ChatGPT API.

Table \ref{tab:summeval_detail} shows that changing prompts result in a significant change in the performance of the human-like automatic evaluation using ChatGPT, especially in terms of system-level correlations. From the current results, these changes do not make it to achieve higher correlations with human judgments, except for a modest improvement in a few cases by adding dimension definitions alone.

\subsection{Comparison with human evaluation}

In terms of accuracy, there is still an overall gap between the current automatic human-like evaluations using ChatGPT compared to human experts. Table \ref{tab:summeval_detail} illustrates that in most cases, the correlation between scores given by a human expert and the average of scores given by human experts is substantially better than ChatGPT at all levels. However, the correlation between ChatGPT and human evaluations (0.889) is already higher than that of a particular human expert (0.843) in terms of system-level correlation of fluency.

For variance and reproducibility, automatic human-like evaluations using ChatGPT are more controllable. It is easy to know from Table \ref{tab:summeval_detail} that the scores of the same samples will not be identical between different human annotators. \citet{belz-etal-2021-reprogen} pointed out that reproducing the manual evaluation was difficult. In contrast, we can make the human-like manual evaluation based on ChatGPT reproducible by setting randomness parameters (e.g., temperature) at decoding time.

In terms of cost, it is cheaper to perform the human-like automatic evaluation. Taking SummEval as an example, in our experiments, the assessment of one summary consumed about 1000 tokens, and it took about $0.002\times 1600=3.2$ USD \footnote{\url{https://openai.com/pricing}} to finish the evaluation on the whole dataset. Assuming that a single annotator spends 5 hours annotating the whole dataset. It costs $12\times 5=60$ USD. It is estimated that the cost of human evaluation is about 10 to 20 times higher than human-like automatic evaluation using ChatGPT.

\begin{table*}[t]

\begin{tabular}{|l|l|}
\hline
\textbf{Description} & \textbf{Example }\\ \hline
\begin{tabular}[c]{@{}l@{}} Refuse to evaluate \end{tabular} & \begin{tabular}[c]{@{}l@{}}I'm sorry, but the proposed summary is not acceptable. It contains \\ inaccurate and irrelevant information, and it is not coherent or \\ fluent.  Please provide a new summary for evaluation.\end{tabular} \\ \hline
Not evaluate as required & \begin{tabular}[c]{@{}l@{}}As an AI language model, I cannot evaluate the quality of \\ summaries as it requires subjective judgment. However, I can \\ provide  feedback on the summary's coherence, fluency, and \\ relevance. The proposed summary is poorly written and {[}......{]}\end{tabular} \\ \hline
Write a new summary & \begin{tabular}[c]{@{}l@{}}As an AI language model, I cannot evaluate the quality of \\ summaries as it requires subjective human judgment. However, \\ I can provide a summary of the article that is more accurate and \\ relevant:\\
\\
Referee Mark Clattenburg has been {[}......{]}\end{tabular} \\ \hline
Continue to write the summary & \begin{tabular}[c]{@{}l@{}}land movement overnight. The mass, which has risen some 50ft \\ above sea level, measures roughly 1,000-1,640ft long, and 100ft \\ wide. {[}......{]}\end{tabular} \\ \hline
\end{tabular}

\caption{Examples of invalid responses generated by ChatGPT on SummEval.}
\label{tab:invalid_cases_exam}
\end{table*}

\subsection{The quality of generated explanations}

We sampled and examined the responses generated by ChatGPT on SummEval, and found the following characteristics of the explanations given by ChatGPT:

ChatGPT sometimes provides scores or labels followed by an explanation, even if it is not explicitly asked to provide the explanation in the prompt. Of course, it is possible to add a request such as "You do not need to explain." to the prompt so that it does not generate an explanation, but the impact of this on the evaluation scores is unknown.

The explanations generated by ChatGPT are generally self-consistent but not necessarily correct. The generated explanations generally coincide with its scoring. For example, Table \ref{tab:expl_cases} shows that ChatGPT and ChatGPT+def both scored low for the faithfulness of the summary, and they both pointed out factual errors in the summary. However, the correctness of these explanations still needs further testing.

The combination of ChatGPT's explanations and scoring can better confirm whether it understands the requirements of the evaluation, for example, the dimension definitions. Without providing dimension definitions (see Figure \ref{fig:likert_scale_detail}), ChatGPT's understanding of fluency and coherence converged. After examining multiple samples we found that its explanations of the scoring of these two dimensions are placed together and the dataset-level correlation between the scoring of these two dimensions is 0.960. ChatGPT is better able to distinguish between these two dimensions when dimension definitions are provided. Its explanations of the scoring of the two dimensions are separated and the dataset-level correlation between the two dimensions drops to 0.843.

\subsection{Invalid responses}

ChatGPT sometimes generates invalid responses, but this fraction is only about 1\% at most (see Table \ref{tab:invalid_cases_prop}). As shown in Table \ref{tab:invalid_cases_exam}, common invalid responses were refusing to evaluate, not evaluating as required, writing a new summary, and continuing to write the summary. The reason why invalid responses are generated needs to be further explored.

\begin{table}[h]
\begin{tabular}{lr}
\hline
 & \multicolumn{1}{l}{Invalid responses} \\ \hline
ChatGPT & 0.0000 \\
ChatGPT+def & 0.0003 \\
ChatGPT+def+ins & 0.0106 \\
ChatGPT+sys\_prompt & 0.0013 \\ \hline
\end{tabular}
\caption{Porportions of invalid responses generated by ChatGPT on SummEval.}
\label{tab:invalid_cases_prop}
\end{table}

\section{Related Work}

There are some concurrent studies using LLMs for human-like NLG evaluation. According to \citet{kocmi2023large}, LLMs are currently the most advanced evaluators of translation quality.  \citet{wang2023chatgpt} tested ChatGPT's ability to be an evaluator on three NLG meta-evaluation datasets. \citet{ji2023exploring} explored the effectiveness of ChatGPT in ranking model-generated content. \citet{luo2023chatgpt} investigated ChatGPT's ability to evaluate factual consistency in summarization.. \citet{liu2023gpteval} utilized ChatGPT and GPT-4 to assess the quality of NLG outputs with chain-of-thoughts.

\section{Conclusion}
From the above experiments using ChatGPT for human-like summarization evaluation, the key findings are as follows:

\begin{itemize}
    \item ChatGPT has the ability to perform summarization evaluation using various human evaluation methods. In some instances, it attains a higher correlation with human judgments than existing evaluation metrics.

    \item The performance of ChatGPT on summarization evaluation is highly dependent on prompt design.

    \item Human-like evaluation with ChatGPT is more cost-effective and reproducible than human evaluation.

    \item The explanation generated by ChatGPT is consistent with its scoring.
\end{itemize}




\bibliographystyle{acl_natbib}

\appendix



\end{document}